\documentclass[10pt,twocolumn,letterpaper]{article}

\usepackage{iccv}
\usepackage{times}
\usepackage{epsfig}
\usepackage{graphicx}
\usepackage{amsmath}
\usepackage{amssymb}
\usepackage[numbers]{natbib}
\usepackage[T1]{fontenc}
\usepackage{multirow}

\usepackage[breaklinks=true,bookmarks=false]{hyperref}

\iccvfinalcopy 

\def\iccvPaperID{****} 

\ificcvfinal\fi

\begin{document}

\title{RAIS: Robust and Accurate Interactive Segmentation\\ via Continual Learning}



\author{Yuying Hao, Yi Liu, Juncai Peng, Haoyi Xiong,  Guowei Chen,\\ Shiyu Tang,  Zeyu Chen, Baohua Lai \\
         Baidu Inc.\\
 {\tt\small \{haoyuying, liuyi22\}@baidu.com}
}

\maketitle
\ificcvfinal\thispagestyle{empty}\fi

\begin{abstract}
Interactive image segmentation aims at segmenting a target region through a way of human-computer interaction. Recent works based on deep learning have achieved excellent performance, while most of them focus on improving the accuracy of the training set and ignore potential improvement on the test set. In the inference phase, they tend to have a good performance on similar domains to the training set, and lack adaptability to domain shift, so they require more user efforts to obtain satisfactory results. In this work, we propose RAIS, a robust and accurate architecture for interactive segmentation with continuous learning, where the model can learn from both train and test data sets. For efficient learning on the test set, we propose a novel optimization strategy to update global and local parameters with a basic segmentation module and adaptation module, respectively. Moreover, we perform extensive experiments on several benchmarks that show our method can handle data distribution shifts and achieves SOTA performance compared with recent interactive segmentation methods. Besides, our method also shows its robustness in the datasets of remote sensing and medical imaging where the data domains are completely different between training and testing.
\end{abstract}

\section{Introduction}
\label{sec:intro}
Deep learning methods have shown superior performance on segmentation tasks~\cite{ocr,li2020improving,liu2021paddleseg}, such as portrait segmentation~\cite{chu2022pp}, satellite image processing~\cite{hua2021semantic}, intelligent driving~\cite{tabelini2021keep}. Unusually, most of them require large-scale annotated images to learn powerful abstraction. However, the cost of manual annotation grows rapidly, as the number of data increases, especially when it comes to pixel-level
segmentation tasks. To improve the efficiency of the annotation process, interactive segmentation appears to be an effective and auxiliary way, which is a semi-automatic method utilizing human-computer interaction. It allows the annotators to provide a small number of interactive information and generates the final segmentation result progressively. Therefore, it can accelerate segmentation annotation while maintaining satisfactory quality. Recently, interactive segmentation has attracted intensive attention in both academia and industry.
\begin{figure*}[t]
	\begin{center}
		\includegraphics[width=0.95\linewidth]{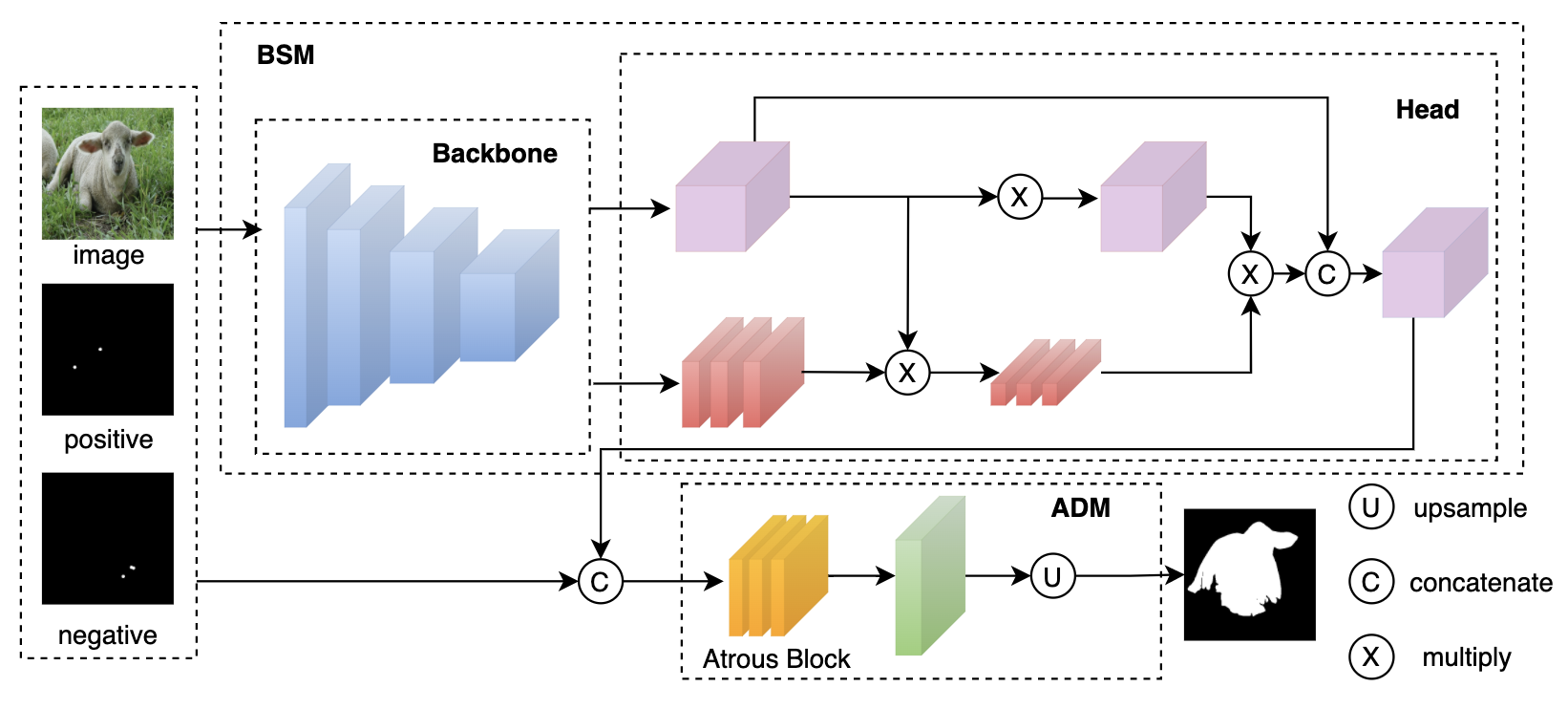}
	\end{center}
	\caption{Overview of the RAIS architecture. It consists of two parts: a basic segmentation module (BSM) and an adaptation module (ADM).}
	\label{fig:structure}

\end{figure*} 
In interactive segmentation, there have been a few types of interactive information, e.g., bounding box~\cite{bbox}, scribbles~\cite{scribbles} or clicks~\cite{ritm,dextr,brs,fca,f-brs}, where their characteristics have been studied well by previous works. Among them, the click-based interactive way is the most widely used, because it provides sufficient region-of-interest information with minimal interaction time. In general, click-based methods usually employ two kinds of user clicks, i.e. positive clicks and negative clicks, which indicate the target region and non-target regions, respectively. In general, most interactive methods~\cite{ritm,dextr,fca} train the model over a training set without updating its parameters at test time. Usually, they do well on the test data similar to the train set. As the difference in data distribution increases, their performance could deteriorate significantly. Accordingly, they require more user clicks to refine the final results, or even they need to be re-trained on the new data, which is increasing annotation costs.

In this work, we propose RAIS, a robust and accurate architecture for interactive segmentation with continuous learning, to address the deterioration problem. In our method, we take interactive segmentation as a continuous adaptation and allow the model to learn from both the train set and test set. For the train set, we use the full-supervision way to update the model parameters like other methods. As for the test set, we propose a weakly-supervised method to refine the model by utilizing the user annotations and intermediate output. Since the user interactions have already provided useful hints of ground truth, the intermediate results are the potential to improve performance for subsequent data. Hence, our model can adapt to the new data distribution gradually, and relieve the impact of the deterioration problem. 
Also, to prevent the model from forgetting previous knowledge when learning new information, we need to minimize unnecessary changes to model parameters at the test time. Therefore, we propose a novel optimization strategy for updating two parts of the model, i.e. basic segmentation module (BSM) and the adaptation module (ADM). BSM retains fundamental knowledge learning from the initial train set, and it would not be updated frequently. ADM learns the residual feature representation between the train set and the test set, and it is updated more frequently to fit new data distribution. Compared to optimizing the overall model directly, our strategy improves the model robustness. With the proposed continuous adaptation strategy, the comprehensive evaluations show our superior performance on well-known benchmarks.

Our contributions are summarised as follows:
\begin{itemize}
\item We propose a robust and accurate architecture for interactive segmentation with continual learning. Through continuous adaptation, the model fits the new data distribution gradually and relieves the deterioration problem of distribution shift.

\item We propose a novel optimization strategy for interactive segmentation tasks at test time. By minimizing unnecessary changes to model parameters, the model prevents from forgetting previous knowledge when learning new information.

\item Extensive experiments demonstrate that our method achieves SOTA performance on several benchmarks. Moreover, our method can improve the robustness of domain changes with fewer user clicks.
\end{itemize}

\section{Related Works}

The interactive segmentation task aims to obtain an accurate mask of an object with minimal user interaction. Interactive information can be clicks~\cite{f-brs,brs,ritm}, bounding boxes~\cite{bbox}, extreme points~\cite{dextr} or scribbles~\cite{scribbles}. Interactive image segmentation has existed for decades. Traditional methods usually optimize energy functions to obtain the object of interest. These methods utilize only low-level features to distinguish foreground and background, so they are not either accurate or robust.

Recently, interactive segmentation based on deep learning has developed rapidly. According to whether the model changes its parameters dynamically on user clicks and test data, these methods can be roughly divided into two types: off-the-shelf methods and on-the-fly methods. 

\textbf{Off-the-shelf methods} optimize the model parameters in the training process and freeze parameters when testing. Xu et al.~\cite{deep1} took clicks as additional input to fine-tune the fcn~\cite{fcn} network. It is the first work introducing deep learning into the interactive image segmentation task. Maninis et al.~\cite{dextr} adopted extreme points to extract the region of interest, they considered that annotating extreme points can reduce annotation time and improve the segmentation accuracy compared with bounding box annotations. Chen et al.~\cite{cdnet} refined the segmentation result by conditional diffusing information of user clicks. They refined the self-attention strategy by adding click and mask information to the affinity matrix and making the model understand user intentions easily. Sofiiuk et al.~\cite{ritm} found that a training set has a great impact on interactive performance. They proposed reviving the iterative training method to obtain the model which can re-correct the existing masks. However, these methods can not adapt to domain changes situation. Most interactive segmentation methods obtain the model on the general scene, such as dog, human, table and so on. If users take these models to annotate the data with significant differences in the training set, the performances are unsatisfactory.

\textbf{On-the-fly methods} dynamically adjust the model parameters or activation information to adapt to new data. Jang et al.~\cite{brs} proposed a back-propagating refinement scheme to ensure that user clicks can be correctly classified. They calculated the gradient of activation on the whole model and cost lots of extra time. f-BRS~\cite{f-brs} improved BRS by updating intermediate features of the model, therefore it saves computation compared with BRS. Kontogianni et al.~\cite{eccv2020} updated the entire model parameters to adapt to domain changes. However, it changes the initial parameters excessively, leading to catastrophic forgetting of previous knowledge. Therefore, the existing methods can not perform well and extract useful information from user clicks and annotated images.

\section{Method}

\subsection{Network Architecture}
\label{3.1}

As shown in Fig.~\ref{fig:structure}, the architecture consists of two parts: a basic segmentation module (BSM) and an adaptation module (ADM). BSM retains fundamental knowledge learning from the initial train set, and it would not be updated frequently. ADM learns the residual feature representation between the train set and the test set, and it is updated more frequently. In this work, we directly use RITM~\cite{ritm} as our BSM, which is consisting of HRNet-18~\cite{hrnet} and OCRNet~\cite{ocr} because of their excellent segmentation performance. For ADM, we 
design an efficient architecture with three atrous convolution blocks and a simple convolution block. The atrous convolutions can expand the receptive field on the high-resolution features, which can improve the details of the low-level feature.

In the training phase, firstly, we only train the BSM part following the criteria~\cite{ritm} without ADM. Then, we train the whole network but freeze the BSM part. Following~\cite{edgeflow}, we also encode the position of user clicks as binary disks, in which positive and negative disks are the superpixels generated from the user clicks with a radius of 5. The BSM part takes the RGB-channel image, positive and negative binary disks~\cite{edgeflow} as an original input and generates a coarse mask. Then, the adaptation module 
concatenates the original input and the generated coarse mask into a new input to refine the result. In this work, we use normalised focal loss~\cite{edgeflow} to optimize the adaptation module parameters, which can be denoted as: 

\begin{eqnarray}
\label{equ:dt1}
\mathcal{L}_{t}(i, j)=-\frac{1}{\sum_{i, j} (1- p_{i,j})^{\gamma}}(1-p_{i,j})^\gamma\log{p_{i,j}},
\end{eqnarray} 
where $\gamma$ is the hyper-parameter, $p_{i,j}$ is the confidence of segmentation result at the position $(i, j)$, and it can be denoted as:

\begin{eqnarray}
\label{equ:dt2}
p_{i,j}=\begin{cases} p, &y=1\cr 1 - p, & y=0\end{cases},
\end{eqnarray}
where $p$ is the output of the adaptation module, and $y$ is the true label at pixel $(i, j)$. We take ground truth and coarse mask as supervision information to optimize the adaptation module.

\subsection{Continuous Adaptation}
\label{3.2}
\begin{figure*}[tp]
	\setlength{\abovecaptionskip}{0.cm}
	\setlength{\belowcaptionskip}{-0.cm}
	\begin{center}
		\begin{tabular}{cccccc}
			\includegraphics[width=0.14\linewidth]{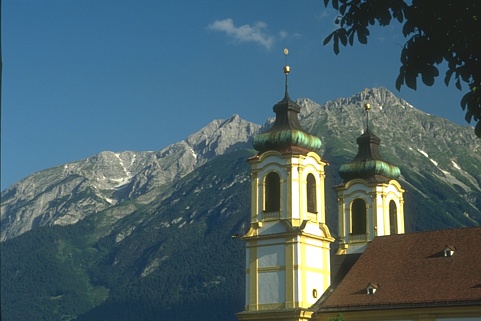}&
			\includegraphics[width=0.14\linewidth]{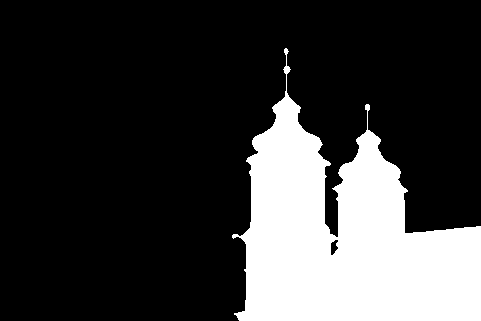}&
			\includegraphics[width=0.14\linewidth]{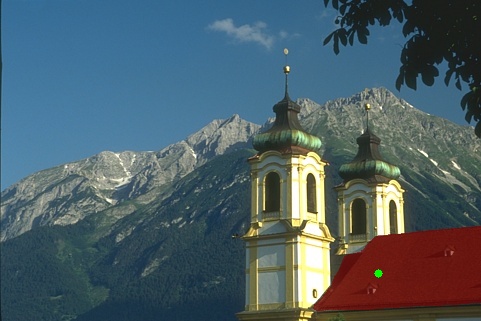}&
			\includegraphics[width=0.14\linewidth]{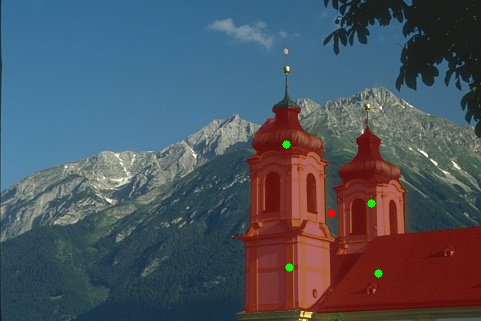}&
			\includegraphics[width=0.14\linewidth]{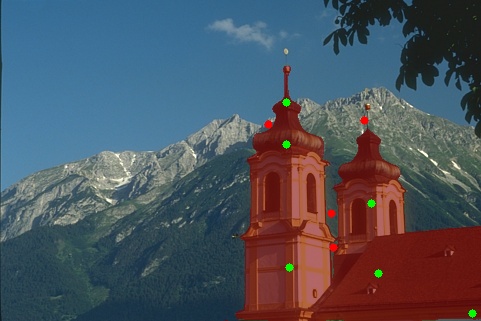}&
			\includegraphics[width=0.14\linewidth]{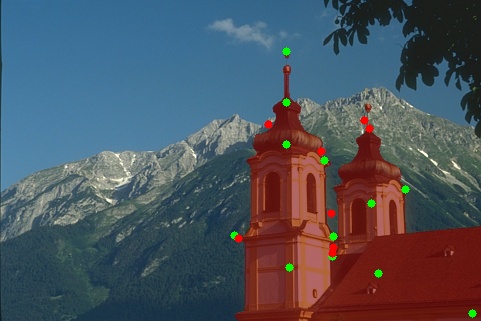} \\
			
			{\scriptsize image} &
			{\scriptsize GT Mask} &
			{\scriptsize IoU=30.54\%} &
			{\scriptsize IoU=89.51\%} &
			{\scriptsize IoU=97.70\%} &
			{\scriptsize IoU=98.10\%} \\

			\includegraphics[width=0.14\linewidth]{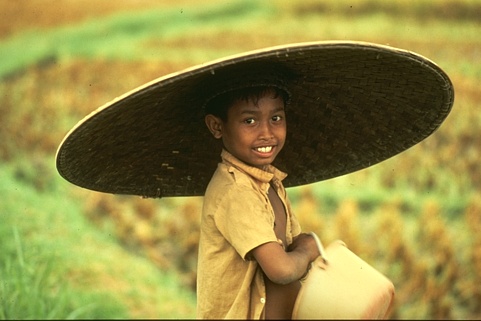}&
			\includegraphics[width=0.14\linewidth]{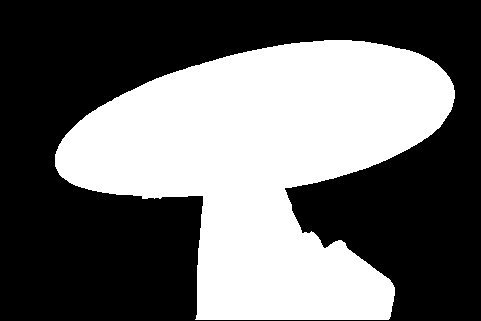}&
			\includegraphics[width=0.14\linewidth]{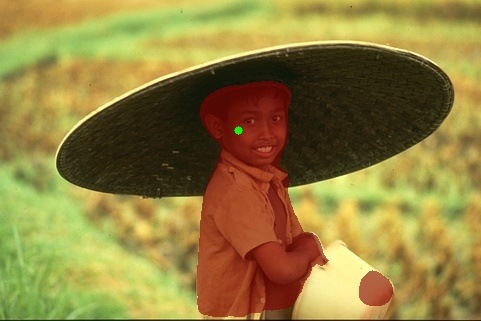}&
			\includegraphics[width=0.14\linewidth]{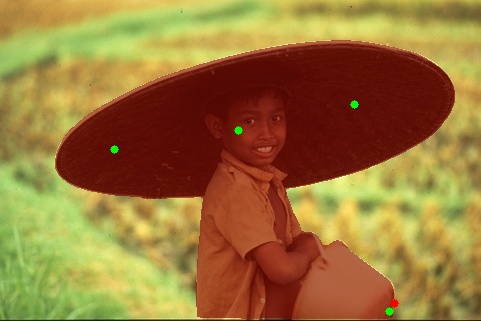}&
			\includegraphics[width=0.14\linewidth]{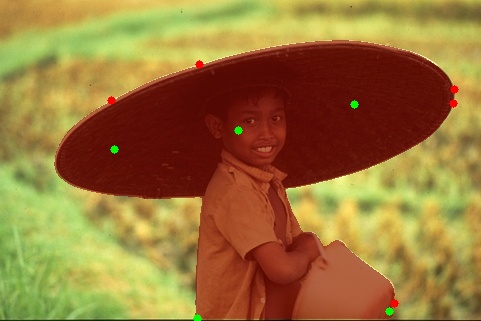}&
			\includegraphics[width=0.14\linewidth]{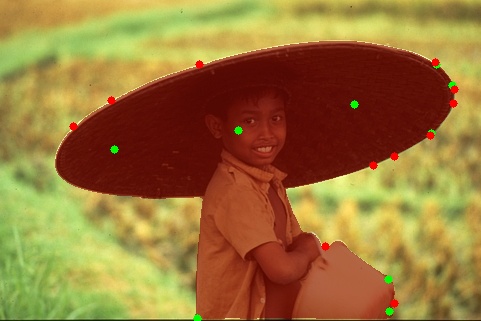} \\
			
			{\scriptsize image} &
			{\scriptsize GT Mask} &
			{\scriptsize IoU=34.33\%} &
			{\scriptsize IoU=98.52\%} &
			{\scriptsize IoU=98.25\%} &
			{\scriptsize IoU=98.05\%} \\
			
			\includegraphics[width=0.14\linewidth]{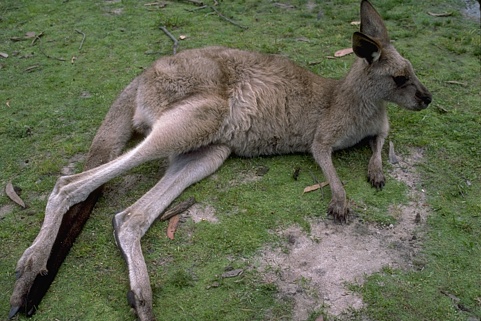}&
			\includegraphics[width=0.14\linewidth]{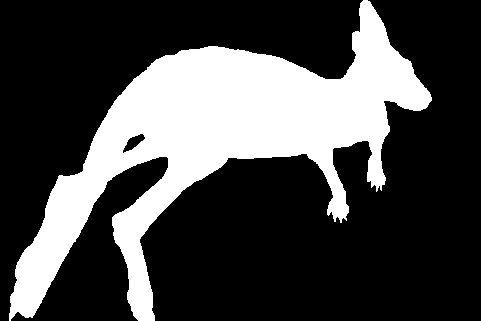}&
			\includegraphics[width=0.14\linewidth]{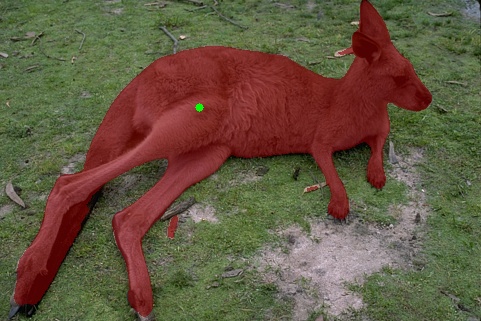}&
			\includegraphics[width=0.14\linewidth]{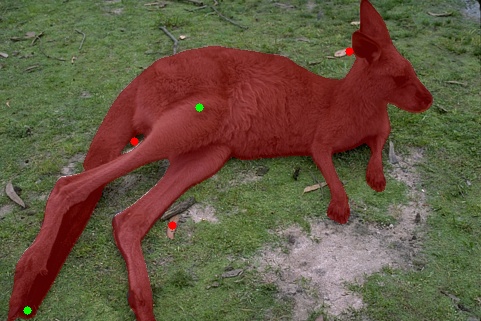}&
			\includegraphics[width=0.14\linewidth]{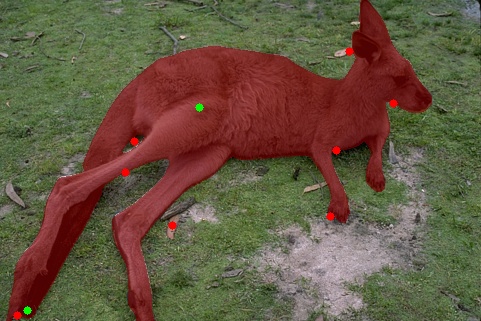}&
			\includegraphics[width=0.14\linewidth]{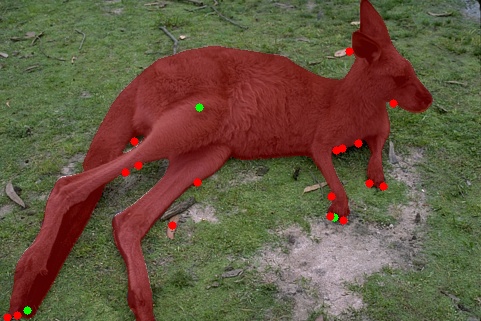} \\
			
			{\scriptsize image} &
			{\scriptsize GT Mask} &
			{\scriptsize IoU=93.64\%} &
			{\scriptsize IoU=95.21\%} &
			{\scriptsize IoU=95.66\%} &
			{\scriptsize IoU=96.18\%} \\
			
			\includegraphics[width=0.14\linewidth]{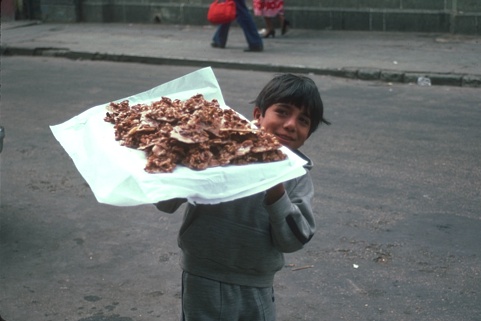}&
			\includegraphics[width=0.14\linewidth]{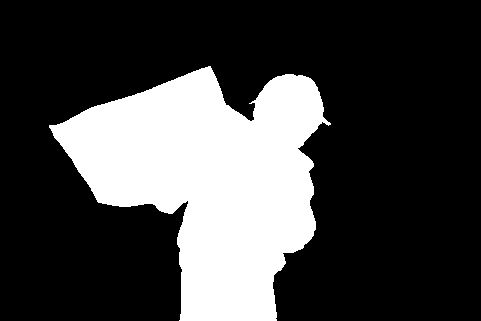}&
			\includegraphics[width=0.14\linewidth]{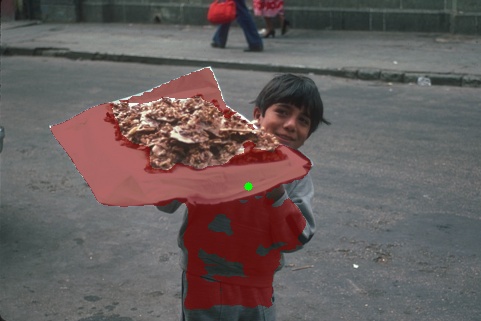}&
			\includegraphics[width=0.14\linewidth]{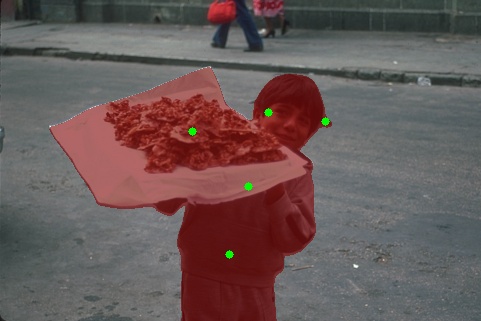}&
			\includegraphics[width=0.14\linewidth]{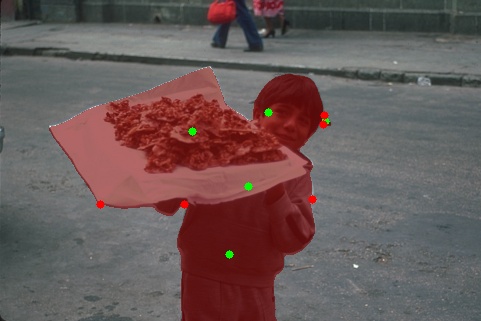}&
			\includegraphics[width=0.14\linewidth]{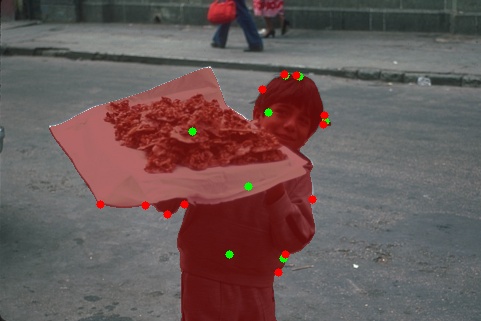} \\
			
			{\scriptsize image} &
			{\scriptsize GT Mask} &
			{\scriptsize IoU=57.94\%} &
			{\scriptsize IoU=97.84\%} &
			{\scriptsize IoU=98.11\%} &
			{\scriptsize IoU=98.29\%} \\			
		\end{tabular}
		
	\end{center}
    \caption{Visualization of our method on Berkeley dataset. The first two columns show images and ground-truth masks. The 3\textsuperscript{rd} to 6\textsuperscript{th} columns visualize the segmentation results with the numbers of clicks for 1, 5, 10, 20, respectively. Green and red dots denote positive and negative clicks, respectively.} 
    \label{fig:show_image1}

\end{figure*}
In general, after the training phase, the model can be used for data annotation directly, i.e., test time. The off-the-shelf methods would not change the model parameters at test time so they ignore potential improvement without learning from the test set. Thus, in this work, we introduce continuous adaptation at test time which allows the model to learn from the test set, and to improve its generalization ability on domain shift tasks. According to the degree of updating parameters at test time, we divide our continuous adaptation method into two modes: local and global optimization. 

\textbf{Local Optimization}: When annotating similar objects or scenes to the training set, the model obtained from the training phase usually shows good performance. Therefore, we only update a small part of model parameters to fit minor distribution shifts, i.e., the ADM part. At test time, the model can not obtain any information from the ground truth. In this work, we use two types of interactive information as supervision for ADM optimization, i.e., user clicks and intermediate annotation results, which provide sufficient hints of ground truth indirectly.

For the user click, we take it as a sparse criterion with the following loss function:

\begin{eqnarray}
\label{equ:dt3}
\mathcal{L}_{s}=\frac{\sum [(1-p)\cdot c_f]^2}{\sum {\bf 1}[c_f=1]} + \frac{\sum [p \cdot c_b]^2}{\sum {\bf 1}[c_b=1]} ~,
\end{eqnarray} 
where $p$ represents the output of adaptation module. $c_f$ denotes the positive disks, and it is a matrix with value \{0, 1\}. $c_b$ is the negative disks. ${\bf 1}$ is an indicator function to calculate the number of pixels with a value of 1.

For the intermediate annotation result, we assume that it tends to be similar to the corresponding ground truth progressively after a few clicks. Then we take it as a dense criterion, and use the same normalize focal loss as Eq.\ref{equ:dt1}, which is denoted as $L_d$. Besides, a $\mathcal{L}_{2}$ regularization is taken to minimize the unnecessary changes in model parameters. It can be denoted as:
\begin{eqnarray}
\label{equ:dt5}
\mathcal{L}_{r}=(\theta - \theta^{\prime})^{2}~,
\end{eqnarray} 
Where $\theta$ represents the updated parameters and $\theta^{\prime}$ represents the last modified model parameters. The total weighted loss function can be represented as:
\begin{eqnarray}
\label{equ:dt4}
\mathcal{L}_{t}=\lambda_{1}L_{s} + \lambda_{2}L_{d} + \lambda_{3}L_{r}~.
\end{eqnarray}

\textbf{Global Optimization}: When the domain changes at test time, the model obtained from the training phase can not perform well. Therefore, the model should re-learn from the test set by updating all parameters. However, if we directly update the whole model parameters, it will change intensely and forget fundamental knowledge learned from the previous training process. In this work, we 
control the degree of parameters update in BSM, and mainly optimize the parameters of ADM. Thus, We set a small learning rate for BSM which is only one per cent of the adaptation module. This strategy can improve the segmentation performance on domain changes while avoiding catastrophic forgetting. We adopt the same loss function as the local optimization strategy to optimize the overall parameters, which is shown in Eq.~\ref{equ:dt4}. Our optimization strategy reduces the BSM parameters update and makes the intermediate results serve as the weak ground truth. Therefore, it can provide a correct direction for BSM fine-tuning.

\section{Experiments}

\subsection{Experiment Setting}

\begin{table*}[tp]
	\centering
	\caption{The Comparison results on GrabCut, Berkeley, DAVIS and Pascal VOC.} 
	\label{tab:sota}
	\begin{tabular}{c|c|c|c|c|c|c} 
		\hline
		\multirow{2}{*}{Method} & \multicolumn{2}{c|}{GrabCut~} & Berkeley~     & \multicolumn{2}{c|}{DAVIS~}   & Pascal VOC     \\ 
		\cline{2-7}
		& NoC@85        & NoC@90        & NoC@90        & NoC@85        & NoC@90        & NoC@85         \\ 
		\hline
            \\
		RW~\cite{rw}                     & 11.36         & 13.77         & 14.02         & 16.71         & 18.31         & -              \\
		ESC~\cite{esc}                     & 7.24          & 9.20          & 12.11         & 15.41         & 17.70         & -              \\
		GSC~\cite{esc}                     & 7.10          & 9.12          & 12.57         & 15.35         & 17.52         & -              \\ 
		\hline
		DOS~\cite{deep1}            & -             & 6.04          & 8.65          & -             & -             & 6.88           \\
		LD~\cite{ld}        & 3.20          & 4.79          & -             & 5.05          & 9.57          & -              \\
		RIS-Net~\cite{ris}                 & -             & 5.00          & 6.03          & -             & -             & 5.12           \\
		ITIS~\cite{itis}                    & -             & 5.60          & -             & -             & -             & 3.80           \\
		CAG~\cite{cag}                     & -             & 3.58          & 5.60          & -             & -             & 3.62           \\
		BRS~\cite{brs}                     & 2.60          & 3.60          & 5.08          & 5.58          & 8.24          & -              \\
		FCA~\cite{fca}                 & -             & 2.08          & 3.92          & -             & 7.57          & 2.69           \\
		IA+SA~\cite{eccv2020}                   & -             & 3.07          & 4.94          & 5.16          & -             & 3.18           \\
		f-BRS~\cite{f-brs}                 & 2.50          & 2.98          & 4.34          & 5.39          & 7.81          & -              \\
	    
	    CDNet~\cite{cdnet}               & 2.22          & 2.64         & 3.69   & 5.17   & 6.66   &  -\\
	    EdgeFlow~\cite{edgeflow}              & 1.60          & 1.72          & 2.40   & 4.54   & 5.77   & 2.50 \\
		RITM-B~\cite{ritm}        & 1.54  & 1.68 & 2.6         & 4.70         & 5.98         & 2.57          \\
		RITM-A~\cite{ritm}        & 1.42  & 1.54 & 2.26        & 4.36         & 5.74         & 2.28          \\ 
		\hline
		RAIS-B        & 1.48  & 1.62 & 2.59        & 4.4        & 5.63         & 2.31          \\
		RAIS-A         & \textbf{1.42}  & \textbf{1.54}  & \textbf{2.19}        & \textbf{4.15}         & \textbf{5.32}       & \textbf{2.08}         \\ 
		\hline
	\end{tabular}

\end{table*}

\begin{figure*}
	\begin{center}
		\begin{tabular}{ccc}
			\includegraphics[width=0.33\linewidth]{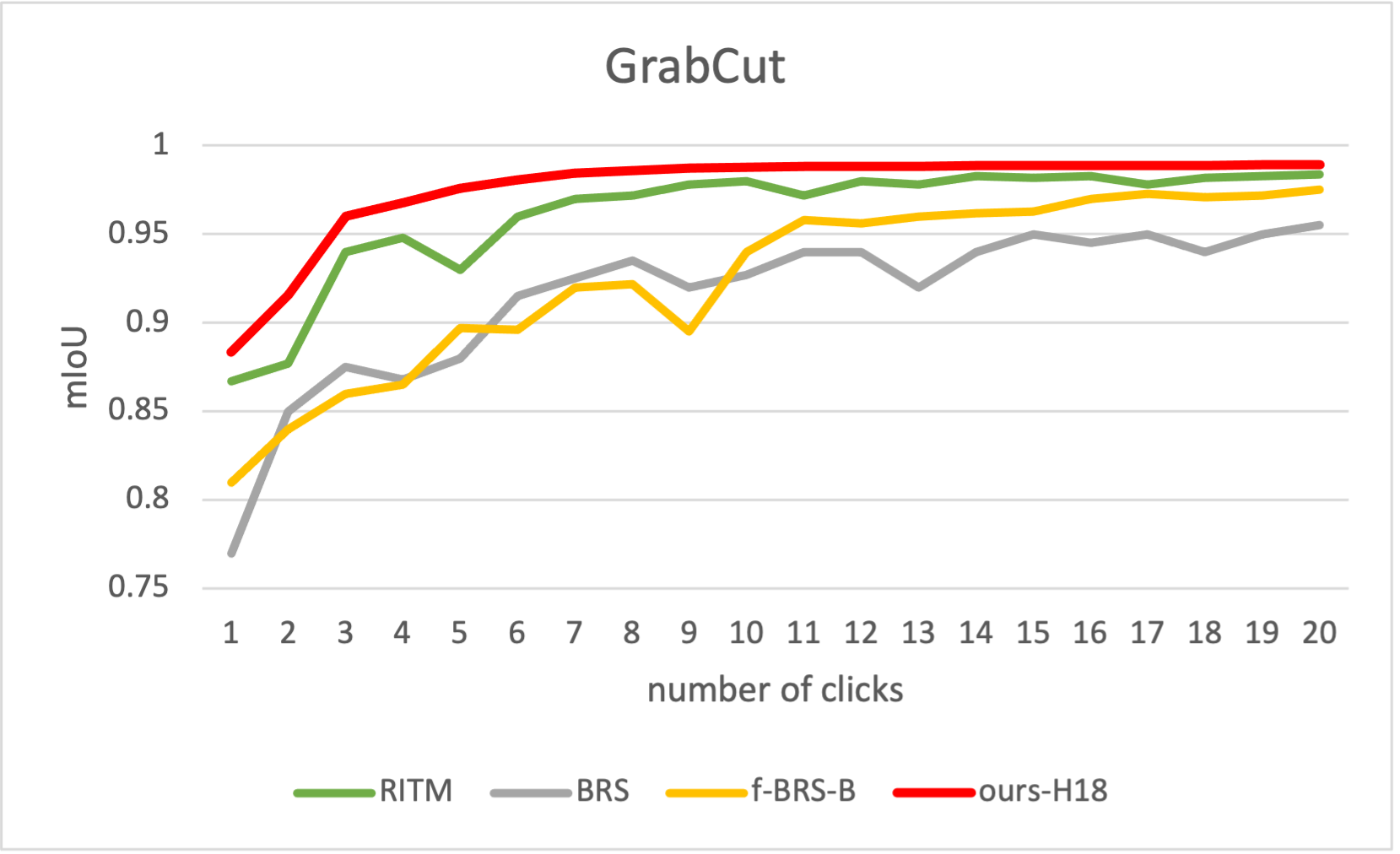}&
			\includegraphics[width=0.33\linewidth]{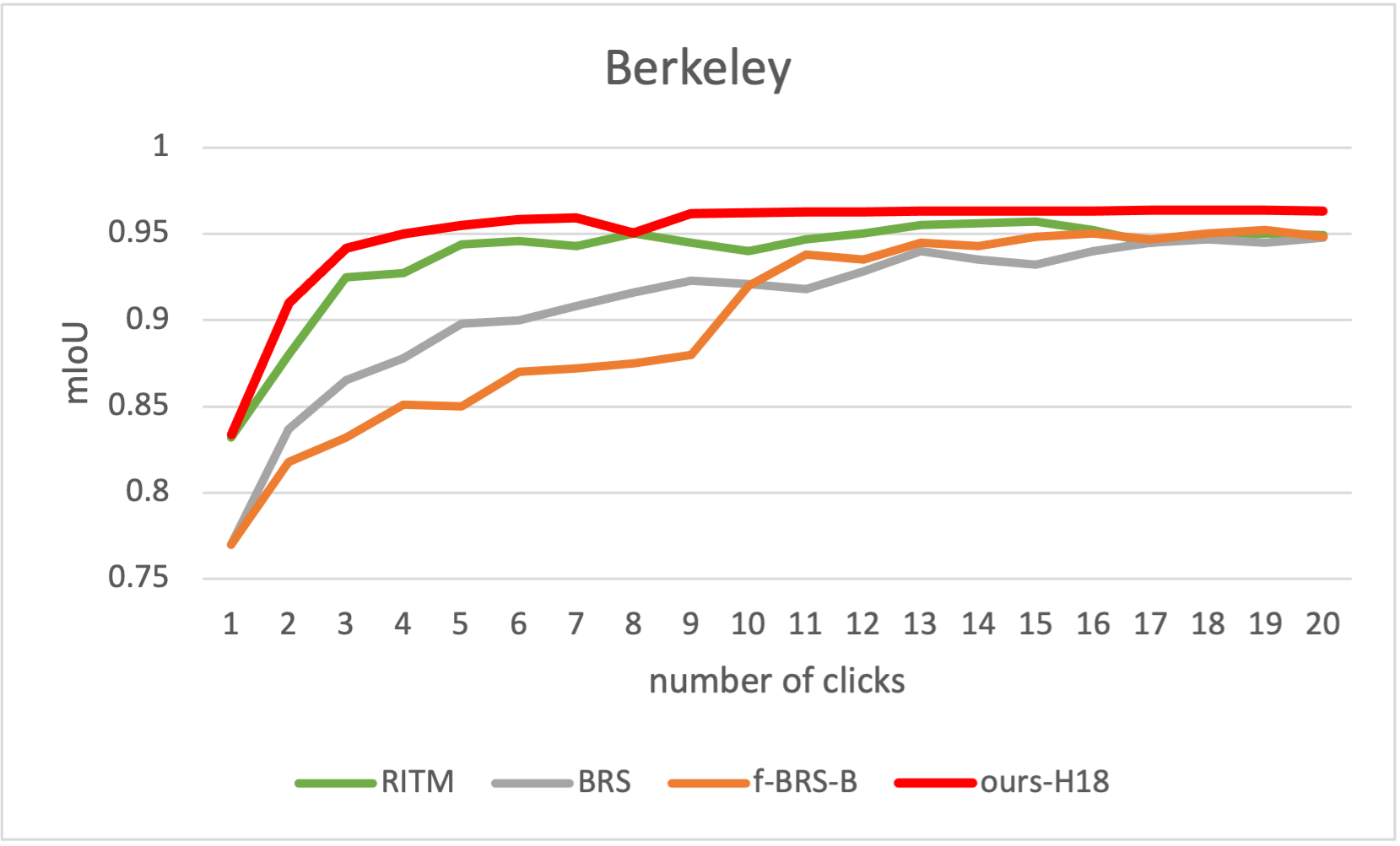} &
			\includegraphics[width=0.33\linewidth]{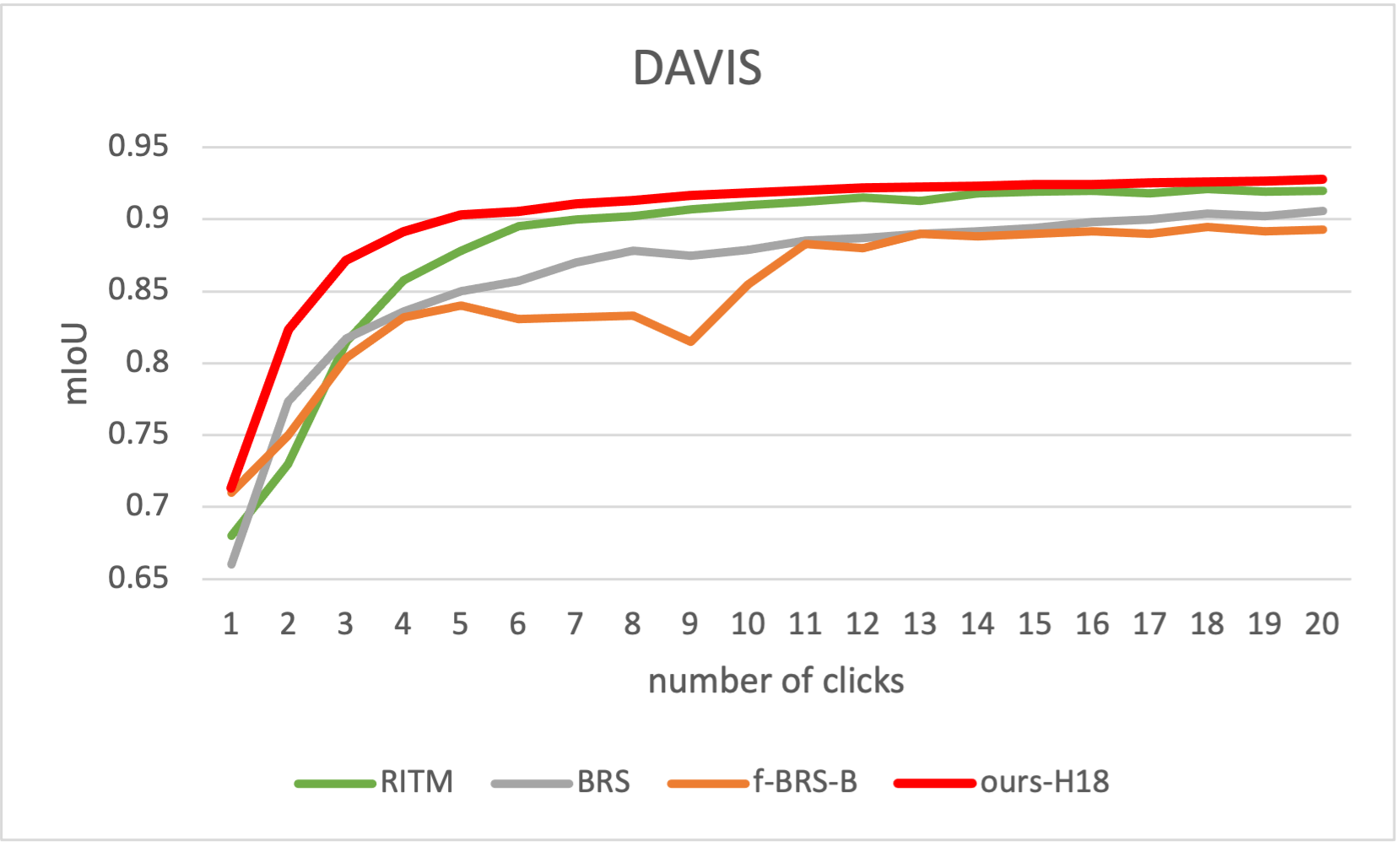} \\
		\end{tabular}
	\end{center}
    \caption{The mean IoU@k for different methods. Please zoom in to see them.} 
    \label{figshow}

\end{figure*}

\textbf{Datasets.} Following RITM~\cite{ritm}, we train our model on LVIS~\cite{lvis} and COCO~\cite{coco}. LVIS has 164K images and 2M instance masks with more than 1K categories. However, due to a large number of categories, it has an obvious long-tail effect that many categories have only a small number of labels. Therefore, we combine 10582 images and corresponding 25832 instance masks from COCO to reduce the long-tail effect. In this work, we evaluate the methods in two scenarios: distribution shift and domain change. 

\begin{itemize}
\item \textit{Distribution shift} indicates that the evaluation set belongs to the same domain as the train set with a minor distribution difference. We evaluate our method on four well-known benchmarks, 1) GrabCut~\cite{grabcut}, which has 50 images and 50 corresponding masks; 2) Berkeley~\cite{berkeley}, which has 96 images and 100 masks; 3) Pascal VOC~\cite{pascalvoc}, which has 1449 images and corresponding masks; 4) DAVIS~\cite{davis}, which contains 345 images and corresponding masks randomly sampled from video object segmentation dataset.

\item \textit{Domain change} indicates that the evaluation set is derived from a different modality compared with the train set. We evaluate our model on remote sensing and medical imaging tasks. For remote sensing, we random sample 200 images and corresponding masks from the mapping challenge and denote it as the Remote-200 dataset. For medical imaging, we random sample 500 images and corresponding masks from Liver Tumor Segmentation Challenge and denote it as LiTS-500.
\end{itemize}
\begin{table*}[tp]
    \vspace{-0.1in}
    \centering
    \caption{Remote sensing and medical dataset adaptation.} 
    \label{tab:remote}
    \setlength{\tabcolsep}{2.5mm}{\begin{tabular}{c|cc|cc|cc} 
        \hline
         \multirow{2}{*}{Model} & \multicolumn{2}{c|}{Method~} & \multicolumn{2}{c|}{Remote~}  & \multicolumn{2}{c}{Medical~}  \\ 
         \cline{2-7}
         & ADM & Optim        & NoC@85        & NoC@90     & NoC@85        & NoC@90           \\ 
        \hline
        \multirow{4}{*}{RAIS-A}    &   -    &     -             & 4.99        & 8.09      & 5.04        & 6.46                \\
            & $\surd$ &      -                    & 5.29         & 8.61     & 5.07         & 6.57         \\
            &   -    & $\surd$                  & 5.30         & 8.01       & 5.06         & 6.46         \\
            & $\surd$        & $\surd$                 & \textbf{4.77}         & \textbf{7.49}     & \textbf{4.77}         & \textbf{6.23}            \\
        \hline
        \multirow{4}{*}{RAIS-B}  & -       &  -                 & 5.30          & 9.05    & 5.68        & 7.26              \\ 
          & $\surd$   &  -                & 5.61        & 9.35            & 5.78            & 7.32      \\
          &  -    & $\surd$                    & 5.43         & 8.94        & 5.37         & 6.76                \\
         & $\surd$   & $\surd$                          & \textbf{4.89}          & \textbf{8.17}        & \textbf{5.23}            & \textbf{6.66}         \\
        \hline
    \end{tabular}}
    \vspace{-0.1in}
\end{table*}

\begin{figure*}[tp]
	\setlength{\abovecaptionskip}{0.cm}
	\setlength{\belowcaptionskip}{-0.cm}
	\begin{center}
		\begin{tabular}{cccc}
			
	        \includegraphics[width=0.23\linewidth]{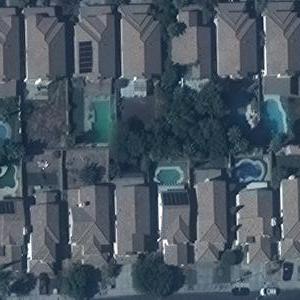}&
			\includegraphics[width=0.23\linewidth]{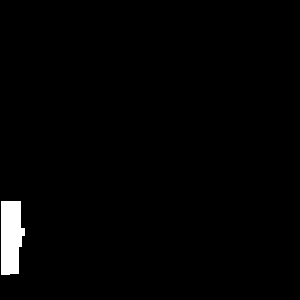}&
			\includegraphics[width=0.23\linewidth]{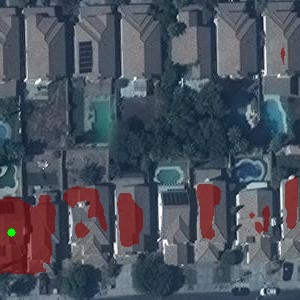}&
			\includegraphics[width=0.23\linewidth]{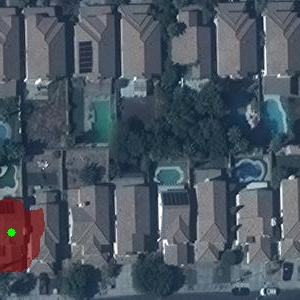} \\
			
			{\scriptsize Image} &
			{\scriptsize GT Mask} &
			{\scriptsize  RITM IoU=18.76\%} &
			{\scriptsize RAIS IoU=51.99\%}  \\	
			
			\includegraphics[width=0.23\linewidth]{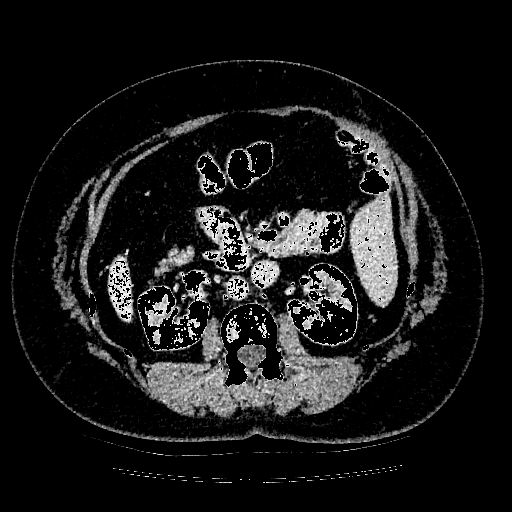}&
			\includegraphics[width=0.23\linewidth]{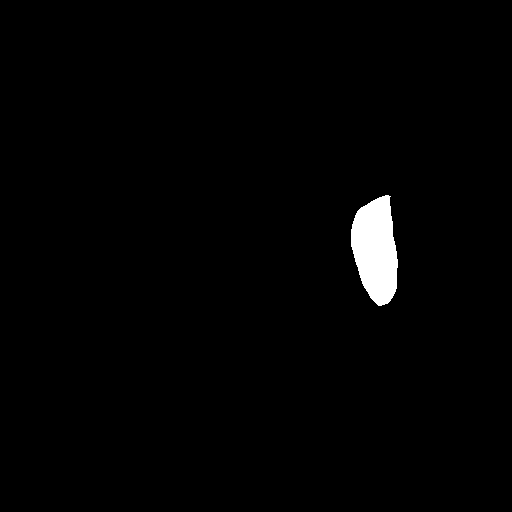}&
			\includegraphics[width=0.23\linewidth]{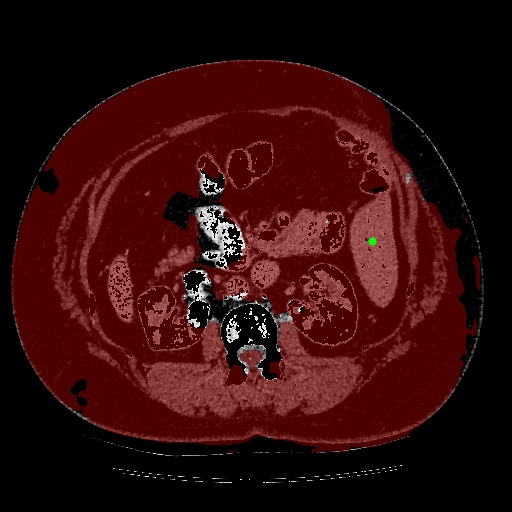}&
			\includegraphics[width=0.23\linewidth]{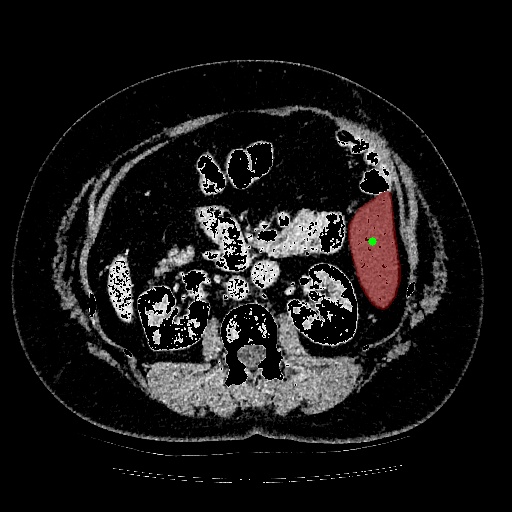} \\
			
			{\scriptsize Image} &
			{\scriptsize GT Mask} &
			{\scriptsize  RITM IoU=2.85\%} &
			{\scriptsize RAIS IoU=84.50\%}  \\

		\end{tabular}
		
	\end{center}
	\vspace{-3mm}
    \caption{The visual comparison on Remote-200 and LiTS-500. Left to right: original images, ground-truth masks, RITM results and RAIS results. The green and red dots denote positive and negative clicks, respectively. Please zoom in to see the details.} 
    \label{fig:remote_medical}
\end{figure*}

\textbf{Evaluation metrics.} We evaluate the average number of clicks (NoC) to achieve the target intersection over the union (IoU) threshold. Following previous works~\cite{ritm,zhang2020interactive, edgeflow}, we also set the target IoU as 85\% and 90\%, which can be denoted as NoC@85 and NoC@90, respectively. The smaller value, the better performance.

\textbf{Implementation details.} At the training phase, the hyperparameters of loss value calculated by ground truth and coarse mask are 1 and 0.4, respectively. For BSM, we adopt RITM~\cite{ritm} H18 and H18S to generate a coarse mask. H18 and H18S are denoted as model `A' and `B', respectively. We train our adaptation module with a learning rate of $5 \times 10^{-4}$. At testing time, the hyperparameters of $\lambda_{1}$, $\lambda_{2}$ and $\lambda_{3}$ are 1, 10 and $5 \times 10^{-3}$, respectively. The learning rate of BSM and ADM is $ 10^{-6}$ and $ 10^{-4}$, respectively. All of our experiments are using EISeg 
at PaddleSeg\footnote{https://github.com/PaddlePaddle/PaddleSeg}~\cite{liu2021paddleseg}.

\subsection{Performance on Distribution Shift}

We evaluate our method and recent SOTA methods on GrabCut~\cite{grabcut}, Berkeley~\cite{berkeley}, DAVIS~\cite{davis} and Pascal VOC~\cite{pascalvoc}. The comparison in Table~\ref{tab:sota} shows the best performance of our method on all benchmarks. The deep learning-based methods have much better performance than traditional ones, e.g., RW, ESC, and GSC. Among modern methods, IA+SA~\cite{eccv2020}, BRS~\cite{brs}, and f-BRS~\cite{f-brs} are on-the-fly methods with the continuous adaption strategy at test time. However, since they do not utilize parameter optimization well, they are worse than SOTA off-the-shelf methods, e.g., EdgeFlow~\cite{edgeflow} and RITM~\cite{ritm}. In comparison, RITM has achieved excellent performance on these benchmarks. We can still improve the interactive performance by utilizing our optimization strategy. As the number of images increases, our method boosts the performance further where it reduces the number of clicks on VOC by 8.8\%.

To better analyze model performance over the number of clicks, we show the mIoU of models on three benchmarks after each user clicks in Fig.~\ref{figshow}. It demonstrates that our method achieves a higher mIoU with a limited number of clicks. When the number of clicks is 5, the mean IoU of our method on GrabCut and Berkeley is above 0.95, and above 0.9 on DAVIS. Also, the curve of our method is more smooth and more stable than other methods, which shows its robustness.

\begin{table*}[t]
\centering
	\caption{Evaluation of catastrophic forgetting on DAVIS.} 
	\label{tab:forget}
    \setlength{\tabcolsep}{2.5mm}{\begin{tabular}{c|c|c|c|c|c}
    \hline
    Strategy & Test  & NoC@85 & Decay   & NoC@90 & Decay   \\ \hline
    \multirow{3}{*}{IASA~\cite{eccv2020}} & DAVIS       & 4.36   & -       &5.74    & - \\
    & Remote + DAVIS   & 4.80   & 10.09\%  & 7.45   & 29.79\% \\
    & Medical+ DAVIS   & 5.04   & 15.60\% & 7.42   & 29.27\% \\
    \hline
    \multirow{3}{*}{RAIS} & DAVIS      & 4.15   & -     & 5.32   & -     \\
    & Remote + DAVIS & 4.42   & 6.50\%  & 5.64   & 6.02\%  \\
    
    & Medical+ DAVIS & 4.56   & 9.88\%  & 5.68   & 6.77\%  \\ \hline
    \end{tabular}}
\end{table*}

\subsection{Performance on Domain Changes}

In this experiment, we train our model on COCO+LVIS~\cite{coco,lvis} and evaluate the performance on Remote-200 and LiTS-500, respectively. In this case, the data domains are completely different between training and testing. 

Table~\ref{tab:remote} shows the adaptation performance of different methods on the two datasets, where ADM denotes whether the model uses the ADM part or not, and Optim denotes whether the model uses the optimization strategy. Note that, if the model does not use either ADM or Optim, it can be regarded as the off-the-shelf method, e.g. RITM. If the model uses Optim, it can be regarded as previous on-the-fly methods. As we can see, our RAIS models with both proposed ADM and Optim show the best performance compared to other methods, which means our method is more robust to domain changes. An interesting observation is that the performance deteriorates with only the ADM part. It means that although the learning ability of the model is becoming stronger with extra components, the adaptability would not get better with it. Thus, A specific adaptation module with a well-designed optimization strategy is necessary for the interactive segmentation model. Furthermore, we show the samples of visual comparison on Remote-200 and LiTS-500 in Figure~\ref{fig:remote_medical}. As we can see, the test images in remote sensing and medical images are much different from the training set, i.e. COO + LVIS. The previous method cannot handle the image annotation with the domain changes, while our method with the well-designed adaptation module and optimization strategy shows the 
much better results.

\subsection{Catastrophic Forgetting}

Catastrophic forgetting refers to the problem that a model forgets previous knowledge after learning from new datasets. On-the-fly methods usually suffer from it, because they update the model parameters arbitrarily during the test time. To show the problem impact, we take the following steps: 1) train the model on COCO+LVIS, 2) test it on remote sense or medical images, and finally, 3) test it on DAVIS. Table~\ref{tab:forget} shows the evaluation results of IASA~\cite{eccv2020} and RAIS, where testing on DAVIS only is the baseline, i.e., without step 2. As we can see, the decay rate of RAIS is much smaller than IASA on both remote sensing and medical images, especially for NoC@90. The evaluation results demonstrate the superior robustness and adaptability of the proposed method that prevents the model from the catastrophic forgetting problem.

\section{Conclusions}

Recent interactive segmentation methods do not fully utilize the ability of continuous adaptation. When predicting on a test dataset with different distributions to the train set, they often suffer from either the deterioration problem or the forgetting problem. In this work, we propose RAIS, a robust and accurate architecture for interactive segmentation with continuous learning. Through continuous adaptation, our model adapts to the new data distribution gradually and relieves the deterioration problem of distribution shift and even domain changes. Also, we propose a novel optimization strategy at test time to minimize unnecessary changes in model parameters, which prevents the model from the forgetting problem. We perform extensive experiments on several benchmarks with both distribution shift and domain change between training and testing. The result shows that our method achieves SOTA performance compared with recent methods.

{\small

\bibliographystyle{ieee_fullname}
}
\end{document}